\documentclass[letterpaper, 10 pt, conference]{ieeeconf}  

\IEEEoverridecommandlockouts                              

\usepackage{graphics} 
\usepackage{epsfig} 
\usepackage{mathptmx} 
\usepackage{times} 
\usepackage{amsmath} 
\usepackage{amssymb}  
\usepackage{multirow}
\usepackage[ruled,vlined]{algorithm2e}
\usepackage{lipsum}
\usepackage{hyperref}
\usepackage{adjustbox}
\usepackage{subfigure}
\usepackage{xcolor}
\hypersetup{
	colorlinks,
	linkcolor={red!50!black},
	citecolor={blue!50!black},
	urlcolor={blue!80!black}
}

\SetAlFnt{\small}
\title{\LARGE \bf
Speed and Separation Monitoring using on-robot Time--of--Flight laser--ranging sensor arrays
}

\author{Shitij Kumar$^{1}$ , Sarthak Arora$^{2}$ , Ferat Sahin$^{3}$
\thanks{$^{1}$ Shitij Kumar *, Ph.D. Candidate Engineering,
        {\tt\small spk4422@rit.edu}}%
\thanks{$^{2}$ Sarthak Arora *, Graduate Student,
        {\tt\small sa9472@rit.edu}}%
\thanks{$^{3}$ Ferat Sahin *, Professor,
        {\tt\small feseee@rit.edu}}
    \thanks{* Department of Electrical and Microelectronics Engineering,
Rochester Institute of Technology, Rochester, NY 14623, USA}%
}

\begin{document}

\maketitle
\thispagestyle{empty}
\pagestyle{empty}

\begin{abstract}

In this paper, a speed and separation monitoring (SSM) based safety controller using three time-of-flight ranging sensor arrays fastened to the robot links, is implemented. Based on the human-robot minimum distance and their relative velocities, a controller output characterized by a modulating robot operation speed is obtained. To avert self-avoidance, a self occlusion detection method is implemented using ray-casting technique to filter out the distance values associated with the robot-self and the restricted robot workspace. For validation, the robot workspace is monitored using a motion capture setup to create a digital twin of the human and robot. This setup is used to compare the safety,performance and productivity of various versions of SSM safety configurations based on minimum distance between human and robot calculated using on-robot Time-of-Flight sensors, motion capture and a 2D scanning lidar.

\end{abstract}





\section{INTRODUCTION}
With the onset of an era of human robot collaboration, a myriad of human robot interaction scenarios can be imagined. To reify these, new and better standards such as \cite{ISOTS15066} have been introduced. Adherence to these standards can mitigate dangers to the human operators whilst enhancing their trust in automation. However, in an industrial context where productivity is also paramount, a system with overly restrictive safety measures may lead to a loss in productivity. Therefore, finding an optimal balance between safety and productivity while still prioritizing the safety is vital.


One of the ways of maintaining the safety of a human operator during human-robot interaction is speed and separation monitoring methodology (SSM) \cite{ISOTS15066}. To achieve SSM, the minimum separation distance and relative velocities between the robot and the human must be determined. This work shows the implementation of a SSM safety configuration using three sensor arrays/rings consisting of eight Time-of-Flight laser-ranging sensors (also known as single-unit solid state lidar(s)) fastened around the robot links as shown in Figure 1, the concept behind which is explained further. The contributions of this work can be listed as:
\begin{list}{\textbullet}{\leftmargin=0.5em}
\item Introduction of a scheme that circumvents the problem of directly computing the true minimum distance between the robot and human whilst encouraging close operation in the robot workspace.
\item Exploration of the viability of intrinsic sensors (on-robot exteroceptive sensors which provide the robot's perspective) for implementing a SSM based safety configuration.
\item An analysis of safety, performance and productivity of an SSM safety configuration by considering different various sensing approaches.
\end{list}

\begin{figure}[t!]
    \centering
    \includegraphics[width=0.4\textwidth,keepaspectratio]{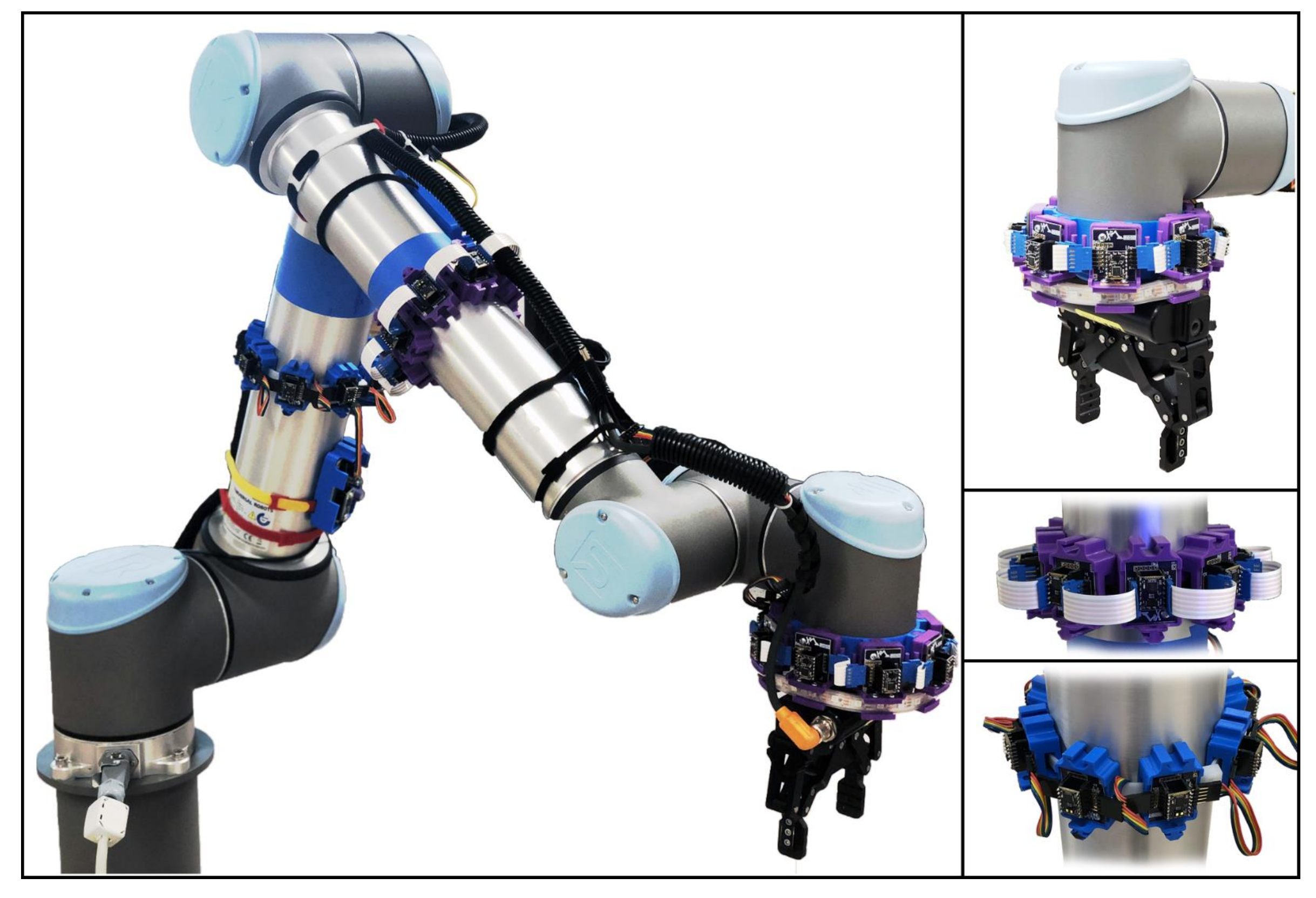}
    \caption{The UR10 robot the with time-of-flight sensor arrays mounted on the robot link centers. Each array has eight single unit lidar(s) This sensing system is used to perform Speed and Separation Monitoring.}
    \vspace{-1.55em}
    \label{fig:my_label}
\end{figure}

In our previous work \cite{kumarDynamicAwarenessIndustrial2018}, a prototype approach was tested with simulated versions of the sensors using only distance between robot and human. This work shows a real world implementation of the same along with considerations of relative human and robot directed speeds while complying to \cite{ISOTS15066}. Using off-robot sensors that are positioned around the robot or on the human operator to maximize workspace area coverage has been the focus of many recent works. Recently, a 2D Lidar was used in conjunction with an IMU based human motion tracking setup \cite{safeeaMinimumDistanceCalculation2019}. Five IMUs were attached to the upper body parts (chest and arms) whereas legs were tracked using a 2D Lidar. Their method of minimum distance calculation was derived from their previous work in \cite{safeeaMinimumDistanceCalculation2017}, where QR factorization was used to find the distances between capsule approximations of human and the robot. In \cite{flaccoDepthSpaceApproach2015}, the authors used RGB-D cameras and proposed a novel approach to compute minimum distances in depth space instead of the cartesian space and also introduced the idea of robot body approximation using few keypoints. We rely on the contribution of the aforementioned and \cite{marvelPerformanceMetricsSpeed2013}, \cite{marvelImplementingSpeedSeparation2017}, where the authors provided metrics for speed and separation monitoring, the work was continued in \cite{marvelImplementingSpeedSeparation2017} where two 2D lidar(s) were used to track the human position with respect to a suspended manipulator. The authors suggested that there was a need to track the human with more precise systems such as motion capture systems. Our work attempts to fill in the gaps by implementing a simpler approach to approximate minimum distances between the human and the robot while drawing a comparison against a complete motion tracking setup where the digital twin of the human is used inside the simulator to calculate an inter-mesh minimum distance between the robot and the human. 

All the approaches mentioned so far have exclusively used proximity or inertial based sensing modalities and approaches, extrinsic to the robot. However, in \cite{schleglVirtualWhiskersHighly2013}, the authors introduced a new type of intrinsic perspective capacitive sensor that encouraged close operation between the human and the robot. In \cite{cerianiOptimalPlacementSpots2013} and \cite{lacevicKinetostaticDangerField2010}, the authors assessed the placement and orientation of IR distance sensors on a robot manipulator and implemented a kineostatic safety assessment algorithm, respectively. Influenced by the aforementioned, our approach employs intrinsic sensors and aims to encourage closer operation. In \cite{lacevicKinetostaticDangerField2010} and \cite{cerianiOptimalPlacementSpots2013}, authors used distance sensors for potential fields and tested the sensors placement on the robot body to examine the work space area coverage.

\section{Methodology}
\label{sec:approach}
\begin{figure}[h!]
\centering
    \includegraphics[width=0.48\textwidth]{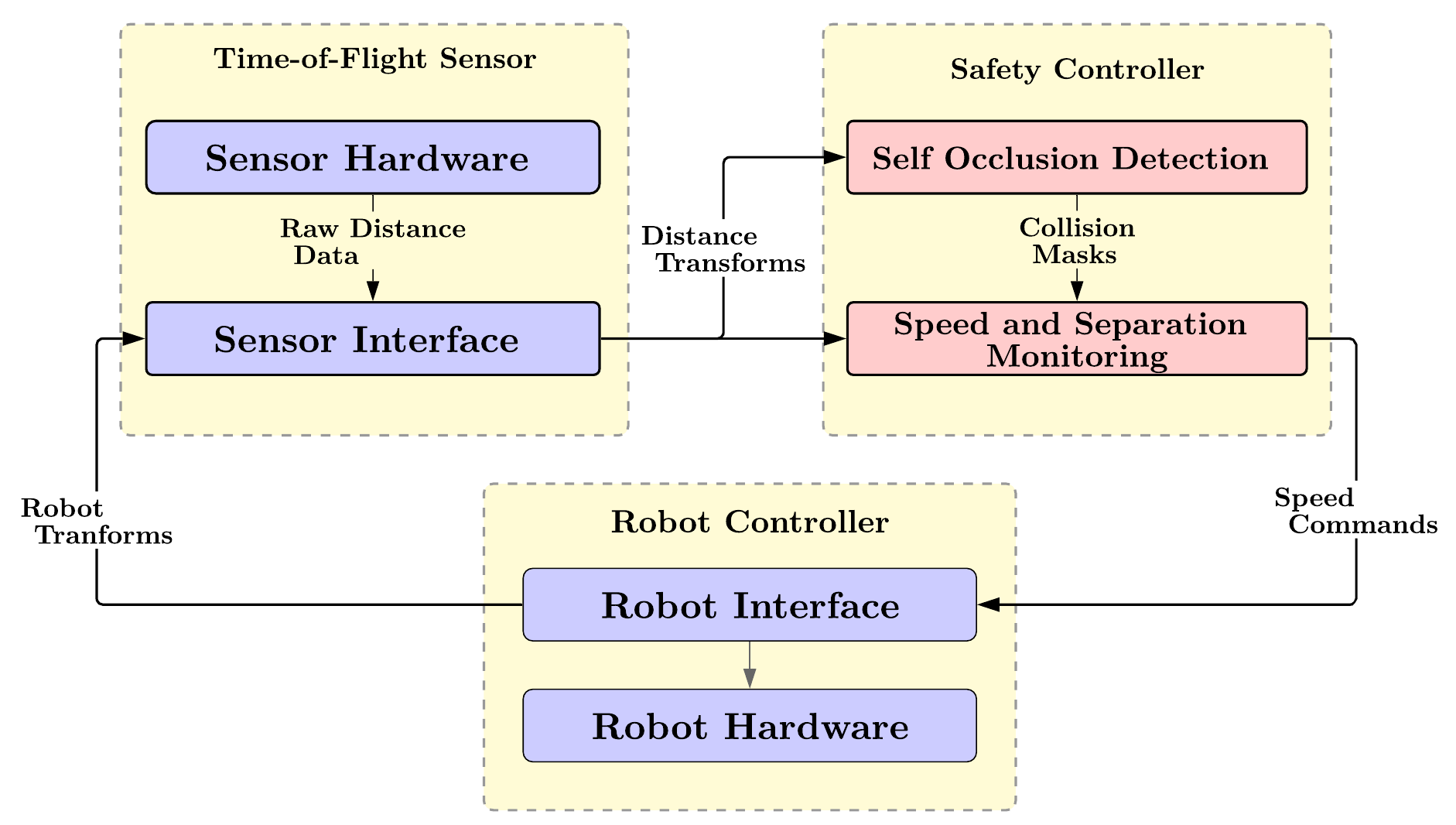}
    \caption{A High-Level Block Diagram representing the proposed setup using TOF sensor arrays. It must be noted that the sensor interface is also responsible for merging the robot kinematic chain with the raw distances provided by the sensors.}
     \label{fig:blockdiag}
\end{figure}

A speed and separation monitoring method using  ToF Sensor arrays  is proposed. The system is implemented as a replacement  of the conventional 2D scanning lidars.
The block diagram of the proposed ToF- Sensor array based  SSM is shown in Figure \ref{fig:blockdiag} . Our approach can be broken down into three components; the SSM formulation for on-robot ToF sensor rings, the minimum distance calculation using the ToF rings and control and communication of the robot.

\subsection{Speed and Separation Monitoring}
 The third collaborative scenario of ``Speed and Separation Monitoring (SSM)" from ISO/TS 15066 \cite{ISOTS15066}, the critical distance $d_C$ (also know as minimum protective distance) at a given current time $t$ can be defined as
\begin{equation}
\label{eq:ssm_linear}
\begin{split}
d_C(t) \geq \underbrace{( V_H(t)T_R + V_H(t)T_S )}_{\text{human distance}} +\underbrace{(V_R(t)T_R)}_{\text{robot reaction distance}} +\\\underbrace{B}_{\text{robot braking distance }} +\underbrace{ (C + Z_s + Z_r)}_{\text{cushioning constant}} 
\end{split}
\end{equation}
In the equation eq.\ref{eq:ssm_linear}, $V_H(t)$ is the speed of human towards the robot, and $V_R(t)$ is the robot speed in direction of the human at current time $t$. $T_R$ is the robot and sensor reaction time to detect a presence in the workspace. $T_S$ is defined as the stopping time and $B$ is the minimum robot braking distance. The terms $C$, $Z_s$ \& $Z_r$ represent the intrusion distance, the operator position uncertainty and the robot pose uncertainty respectively.  Going forward these terms will be combined as a constant $C_{dc}$, \textit{cushioning constant}. 
Eq.\ref{eq:ssm_linear} is used to derive the critical ($d_C$) and reduced ($d_R$) distances for the speed and separation monitoring using the ToF rings.

An SSM safety configuration considers the current separation distance between the human and robot in conjunction with their directed speeds towards each other. This can be  interpreted as the distance between the two closest points on the robot and the human.  While the directed speeds of the robot and the human can be interpreted as their velocity projections on the vector defined by these points .

There are essentially two types of safety distances; critical and reduced speed distance. An SSM configuration that uses these distances to operate the robot at three different speed levels; \textit{pause, slow and normal} is called Trimodal SSM \cite{marvelPerformanceMetricsSpeed2013}, which has been further referred to as TriSSM for brevity. Also, when only the fixed safety thresholds are used (the robot or human directed speeds are not considered), this safety setup is called Tri Modal Separation Monitoring (Tri-SM).

Three different versions of TriSSM  based on the influence of directed human and robot speeds are shown and formulated further. They are :
\begin{itemize}
    \item \textbf{\textit{TriSSM-Vo}} - Tri Modal SSM with relative human/obstacle -robot directed speed ($V_o$).
    \item \textbf{\textit{TriSSM-Vr}} - Tri Modal SSM with only the robot directed speed towards the human/obstacle ($Vr$).
    \item \textbf{\textit{Tri-SM}} - Using fixed distance thresholds on the minimum distance reported by the sensors ($SM$). 
\end{itemize}

As stated above, the outcome of Tri-SSM is to reduce the speed of the end-effector without deviating from the intended trajectory of the task the robot has been programmed to do. Most of the controllers have the option to reduce the operating speed of the robot by a fraction. Here we define the state of the robot as $\psi  \in  \lbrace  \psi_{stop}=0,\psi_{reduced}=1,\psi_{normal}=2 \rbrace$. This determines the mode of operation for the robot performing the task.  
Let $\Dot{x_e}$ be the linear velocity of the Tool Control Point (TCP) (end effector), $(q,\Dot{q})$ be the joint angles and joint velocities of the robot, then the relation can be represented using a Jacobian $J_e$  \cite{khatibUnifiedApproachMotion1987} as :  
\begin{subequations}
\begin{equation}
\label{eq:jacobian}
   \Dot{x_e} =  J_e  \Dot{q}
\end{equation}
\begin{equation}
\label{eq:jacobian_FT}
    \tau = J_e{^T}  F_e 
\end{equation}
\end{subequations}

Given a scalar fraction $\rho$ of the operational speed.
\begin{equation}
\label{eq:jacobian_speed_fraction}
      \Dot{x_e} =  J_e  \Dot{q} \longrightarrow \rho \Dot{x_e} =  J_e  (\rho  \Dot{q})
\end{equation}
\begin{equation*}
\text{where }\rho= 
\begin{cases}
    0,& \text{if } \psi=\psi_{stop}\\
    0.5,& \text{if } \psi=\psi_{reduced} \\
    1, & \text{if } \psi = \psi_{normal}
\end{cases}
\end{equation*}

In order to determine $\psi$, the required computations are formulated hereon.

\begin{figure}[tb]
    \centering
		\includegraphics[width=0.8\linewidth,keepaspectratio]{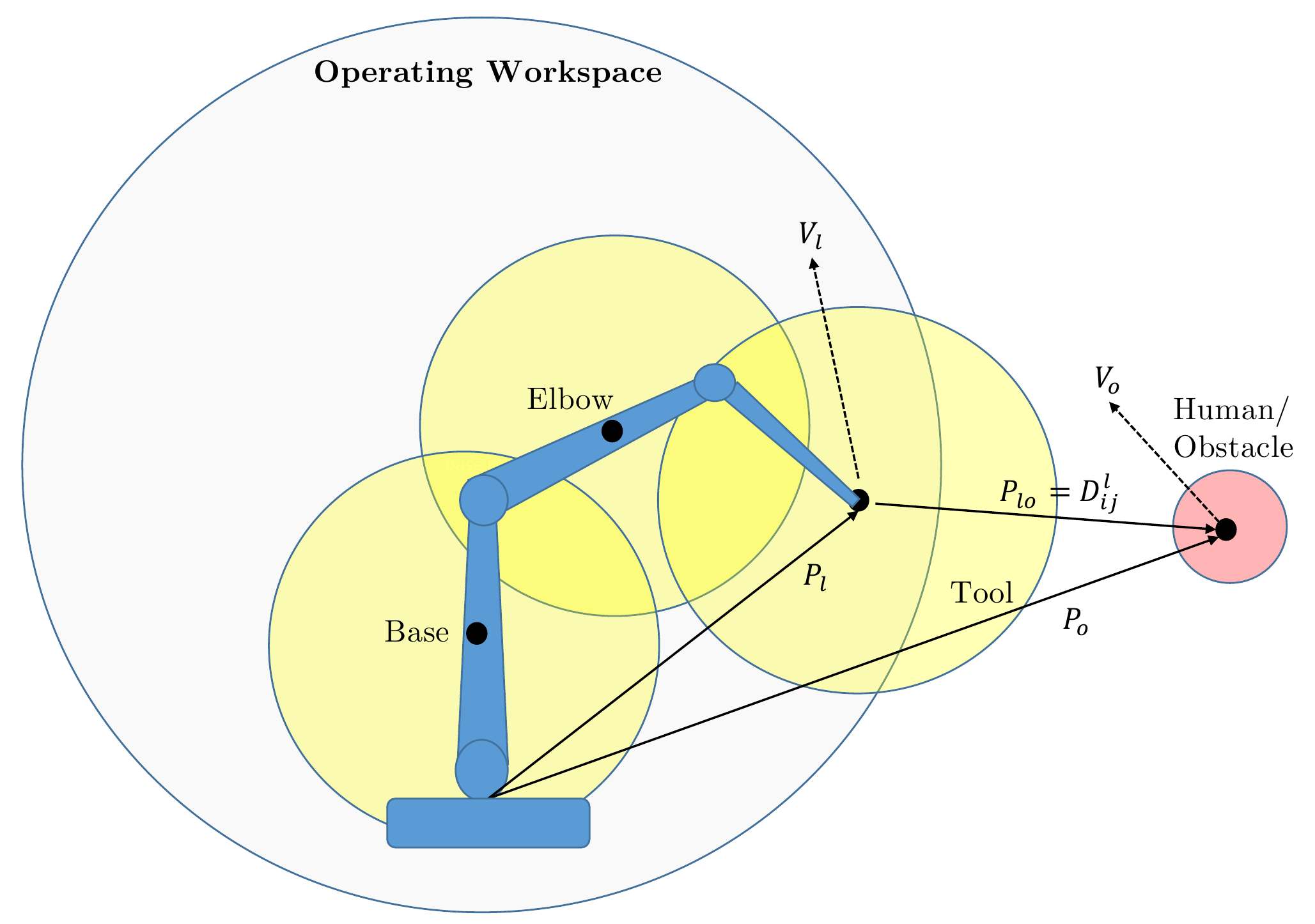}
\caption{A schematic of robot workspace and the detection areas of rings mounted on the robot links. The minimum distance vector $\Vec{^{i}P_{lo}} \equiv \Vec{D_i}$ is represented along with the robot link ($V_l$) and human/obstacle ($V_o$) velocities.}
\label{fig:robot_workspace}
\end{figure}


\subsubsection{\textbf{Minimum Distance Vector}}
\label{sec:minimum_distance_vec}
The `ground truth' minimum distance vector $\vec{D_{gt}}$ can be defined as a`mesh-mesh' minimum distance \cite{rohmerVREPVersatileScalable2013} between the robot and the human/obstacle. 
To overcome direct mesh-mesh comparison(s) and to limit the search space of the computation,  an approximation is made to measure the distance between the two entities where the robot is represented by  using only the centers of the ellipsoidal approximations of the robot links (Figure \ref{fig:sensor_locations}) \cite{flaccoDepthSpaceApproach2015} \cite{safeeaMinimumDistanceCalculation2017}. This can be formulated by defining the 6 Degree-of-Freedom (DOF) robot as a 3-DOF robot with links $\lbrace base,elbow,tool\rbrace$ as shown in Figure \ref{fig:robot_workspace} and explained in \cite{kumarDynamicAwarenessIndustrial2018}.

$\Vec{^{i}P_l}(q)$ and $\vec{P_o}$ are position vectors in the world frame representing the positions of the center of the link $l_i$ ( $i \in \lbrace base,elbow,tool \rbrace$ ) given robot pose $q$, and the human/obstacle, respectively. The approximated minimum distance vector, \textit{Ideal} distance vector $\Vec{D_{ideal}} \approx \Vec{D_{gt}}$  can be defined as :

\begin{equation}
\label{eq:idealeq}
 \forall(i) \in \lbrace base,elbow,tool \rbrace, \Vec{D_{ideal}} = \min (\Vec{^{i}P_{lo}})
\end{equation}


\subsubsection{\textbf{Directed Speeds of Human/Obstacle and Robot}}
\label{sec:directed speed}
As seen in Figure \ref{fig:robot_workspace}, $\Vec{Vo}$ is the detected human/obstacle velocity and $\Vec{^{i}V_l}$ is the link velocity of link $l_i$. As relative directed speeds are calculated w.r.t. all links, $^{i}V_l, ^{i}P_{lo}$, are changed to $V_l,P_{lo}$ for ease of notation. The total directed speed of the human/obstacle towards the center of the robot link $l_i$ is defined as as $k_o(t)$, and the directed speed of the robot alone as $k_l(t)$ at a given time $t$. These can be calculated by projecting the relative velocity of human/obstacle and robot link ($\Vec{V_{lo}}$) and the robot link velocity $\Vec{V_l}$ onto the minimum distance vector $\Vec{P_{lo}}$ respectively. 
\begin{subequations}
\begin{equation}
    k_o= (\Vec{V_l}-\Vec{V_o}) \bullet \vec{P_{lo}} 
\end{equation}
\begin{equation}
    k_l= \Vec{V_l} \bullet \vec{P_{lo}}
\end{equation}
\begin{equation*}
    \text{ where } \bullet \text{\textit{ is a dot-product.}}  
\end{equation*}

\end{subequations}


\subsubsection{\textbf{Critical and Reduced Distances}}
\label{sec:critical_n_reduce_distances}
\begin{figure}[tb]
			\centering
		\includegraphics[width=0.6\linewidth,keepaspectratio]{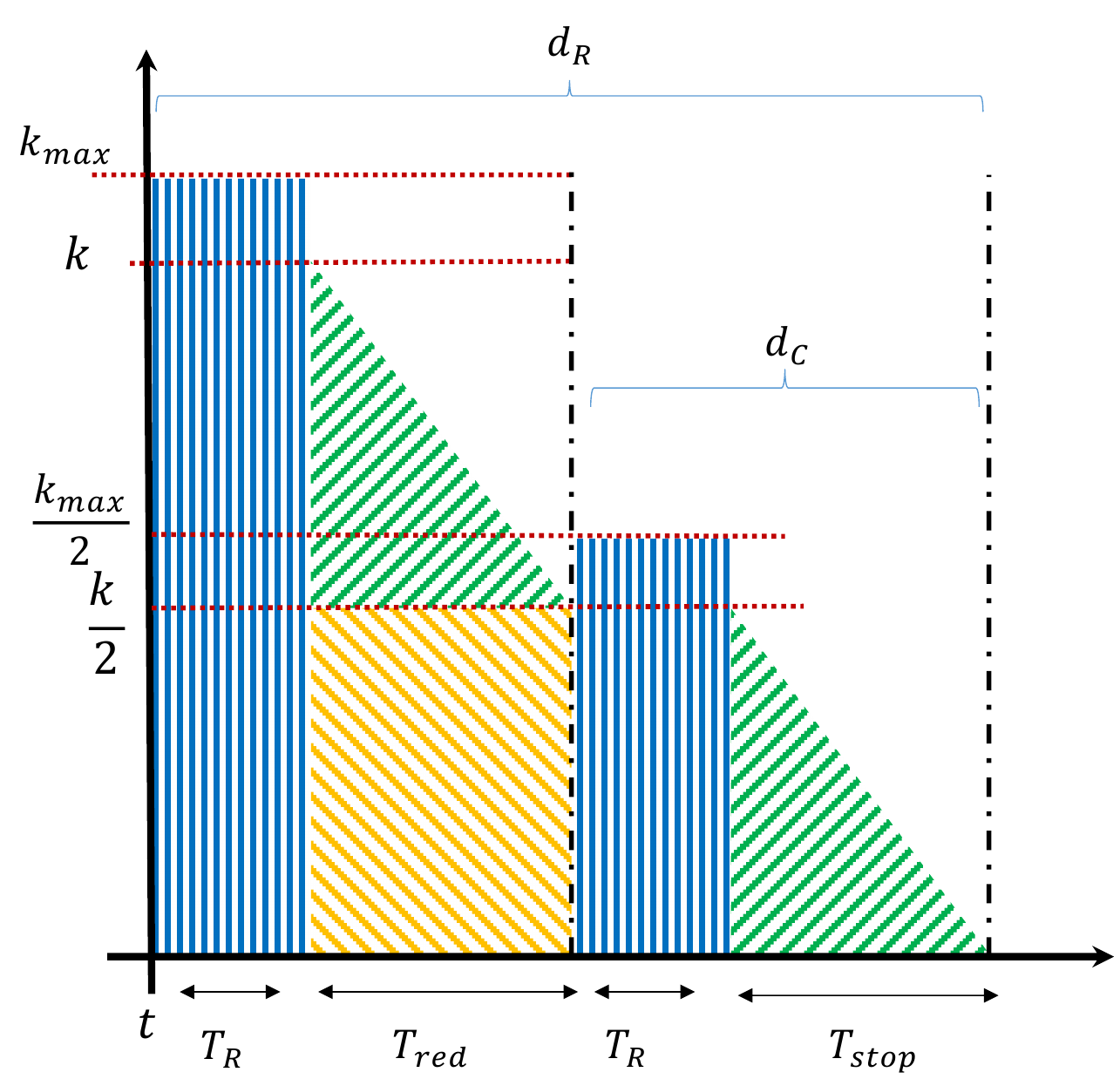}
\caption{(a) Graphs representing the influence of directed speeds of robot and human/obstacle in calculation of safety distance in TriSSM-Vo and TriSSM-Vr configuration.}
	\label{fig:TriSSM_AreaCurve}
\end{figure}

The directed speed influence on the critical $d_C$ and reduced distance $d_R$ can be calculated by estimating the area under the curves of graph shown in Figure \ref{fig:TriSSM_AreaCurve}. $T_{stop}$, $T_{red}$ can be defined as the time taken for the robot to move at half the operating speed.  Let $k \in  \lbrace k_o,k_l \rbrace $ and the max values be $k_{max} \in \lbrace k_{omax},k_{lmax}\rbrace$. Then the final formulation of $d_R(t)$ and $d_C(t)$ at time $(t)$ taking into account the cushioning constant ($C_{dC}$) and minimum braking distance (essentially the time taken by the robot to come to a stop) can be defined as shown in Eq. \ref{eq:reduce_distance} and Eq. \ref{eq:critical_distance}. 

\begin{subequations}
\begin{equation}
    \label{eq:reduce_distance}
d_R(t) = (k_{max} T_R) + k(t) (\frac{3}{4}T_{red}) + R_{buffer}(\psi(t))+ d_C(t)
\end{equation}
\begin{equation*}
    \label{eq:reduced buffer}
    \text{ where } R_{buffer}(\psi(t))= (\psi_{normal} - \psi(t))(\frac{V_{lmax}*T_{red}}{4}) 
\text{ and }
\end{equation*}

\begin{equation*}
d_C(t) = (\frac{k_{max}T_R}{2} )+  (\frac{k(t) T_{stop}}{2}) + \max (B_{min},\frac{\left\lVert V_l(t)\right\rVert T_{stop}}{2} ) 
\end{equation*}
\begin{equation}
\label{eq:critical_distance}
+ C_{dC}  
\end{equation}
\begin{equation*}
    \label{eq:cushioning constant}
\text{ where } C_{dC} = C + Z_s + Z_r
\end{equation*}
\end{subequations}
$R_{buffer}$ is a recovery distance to avoid sudden increase in robot operation speeds when transitioning from $\lbrace \psi_{stop},\psi_{reduced}\rbrace$ to $\psi_{normal}$.
The influence of directed speeds of robot and human/obstacle in calculation of safety distance $(d_C,d_R)$ for a TriSSM-Vo configuration is when $k = k_o$, and $k=k_l$ for TriSSM-Vr. In the Tri-SM configuration the values of safety distances $(d_C,d_R)$ are fixed constants. 


\subsubsection{\textbf{Distance Safety Index}}
 A Distance Safety Index (DSI) for each link $l_i$ is defined as :
 \begin{subequations}
\begin{equation}
	{dsi_i} = \frac{1}{\lVert\Vec{^{i}P_{lo}}\rVert ^2} 
\end{equation}

\begin{equation}
	\begin{split}
	{DSI}_i(t) = \frac{{dsi}_i(t)}{dsi_{max}} \\
	\text{ where } dsi_{max}= (\frac{1}{C_{dC}})^2
	\end{split}
\end{equation}
\begin{equation}
        DSI_C(t) = (\frac{C_{dC}}{d_C(t)})^2 \text{  and  } DSI_R(t) = (\frac{C_{dC}}{d_R(t)})^2
\end{equation}
\end{subequations}
where ${DSI}_k $ is the normalized distance safety index calculated by dividing ${dsi}_k$ by the maximum allowable $dsi_{max}$ i.e. the $dsi$ obtained by the minimum allowable separation distance threshold, i.e. the cushioning constant $C_{dC}$.
The thresholds $\lbrace DSI_C(t), DSI_R(t) \rbrace$ at any given time $t$ are calculated based on critical $d_C$ and reduced $d_R$ distances.

\begin{algorithm}[!h]
\SetAlgoLined
\KwData{distance safety index \& thresholds from ring $i$ \\ $DSI=(DSI_{i}(t))$, $thresholds =(DSI_{{C}_{i}}(t),DSI_{{R}_{i}}(t))$}
\KwResult{$\psi_i$, robot state according to ring i}
read $DSI_{last} = DSI_{i}(t-\Delta t)$ \;
	set hyper-parameters $\alpha_I =0.3,\alpha_D=0.8 $ \;
	//exponential filter for smoothening $DSI$// \\
	\eIf{$DSI  < DSI_{last}$}{
		$\alpha = \alpha_I$ \;
	}{
	    \eIf{$DSI \geq DSI_C$ }{
	    $\alpha =1.0$ \;
	}{
    		$\alpha = \alpha_D$ \;
	}
	}
	$\Hat{DSI} = \alpha DSI + (1 - \alpha)DSI_{last} $ \;
	$ DSI_{last} = \Hat{DSI} $ \;
	\eIf{$ \hat{DSI} \leq DSI_R$ }{
	    $\psi_i = \psi_{normal}$}
	{ \eIf{$ DSI_R < \hat{DSI} < DSI_C $}{
	    $\psi_i = \psi_{reduced}$
	    }{
	    $\psi_i = \psi_{stop}$
	    }
	}
 \caption{Distance Safety Index Algorithm}
 \label{algo:si}
\end{algorithm}

Using the Algorithm \ref{algo:si} the robot state is determined. The overall robot state $\psi = \min (\psi_i : i = {base,elbow,tool} )$ is the minimum of the states reported by the ToF rings i.e. the highest danger presented by the ring is the state of the robot.

\subsection{Time-of-Flight Sensor Array: ToF Ring}
\label{sec:TOFSesnorArray}

\begin{figure}[h]
        \centering
		\includegraphics[ width=0.6\linewidth,keepaspectratio]{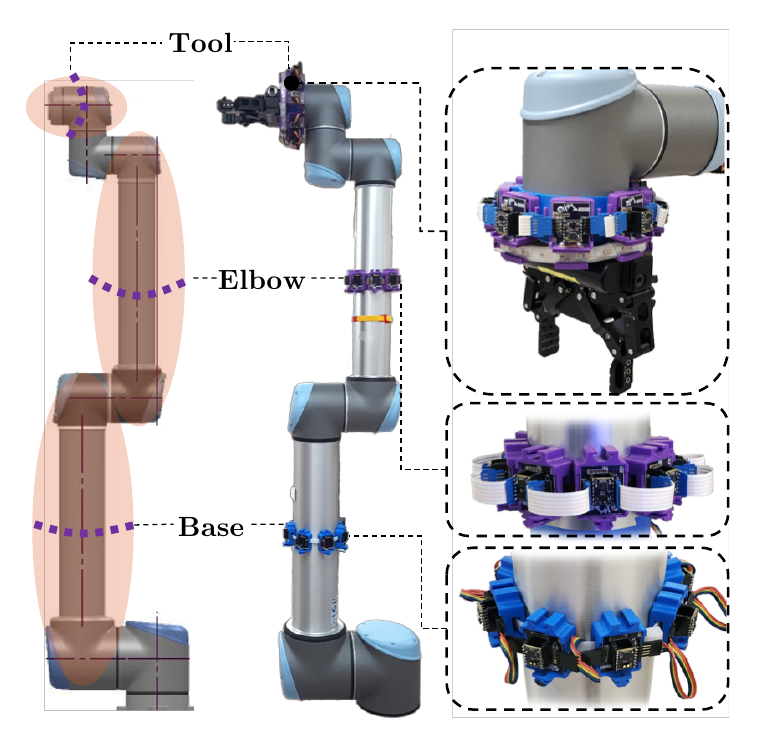}

\caption{A Geometrical Approximation of 6 DOF robot, here the UR 10. Base, Elbow and Tool-end effector for 3 major links of the robot.TOF Ring prototypes mounted on UR10 robot.}
	\label{fig:sensor_locations}
\end{figure}

For obtaining the ideal minimum distance Eq. \ref{eq:idealeq}, an array of eight time-of-flight single unit lidar(s) (\textit{VL53L1X}) is used. The rings are mounted on the the links of the UR-10 robot as shown in Figure \ref{fig:sensor_locations}. The sensors are capable of reading upto $1.5m$, within a field-of-view (FOV) of $25\deg$. The best operating range as per the  from $0.03m -1.31m$ . Unlike \cite{cerianiOptimalPlacementSpots2013}, this placement was chosen to implement the simplest form of minimum distance calculation using a geometric approximation of the links of the robot. Due to the sparsity between the sensors in each sensor array, an approximation, $\Vec{D_i} \approx \Vec{D_{ideal}}$ is assumed.


\subsubsection{\textbf{Minimum Distance Calculation using the TOF rings}}
\label{sec:Minimum Distance Calc TOFRings}
The sensor interface reads data from each sensor ring and combines it with the robot position to determine the minimum distance vector  w.r.t. to each link.
\begin{figure}[h]
		\centering
		\includegraphics[width =0.9\linewidth,keepaspectratio]{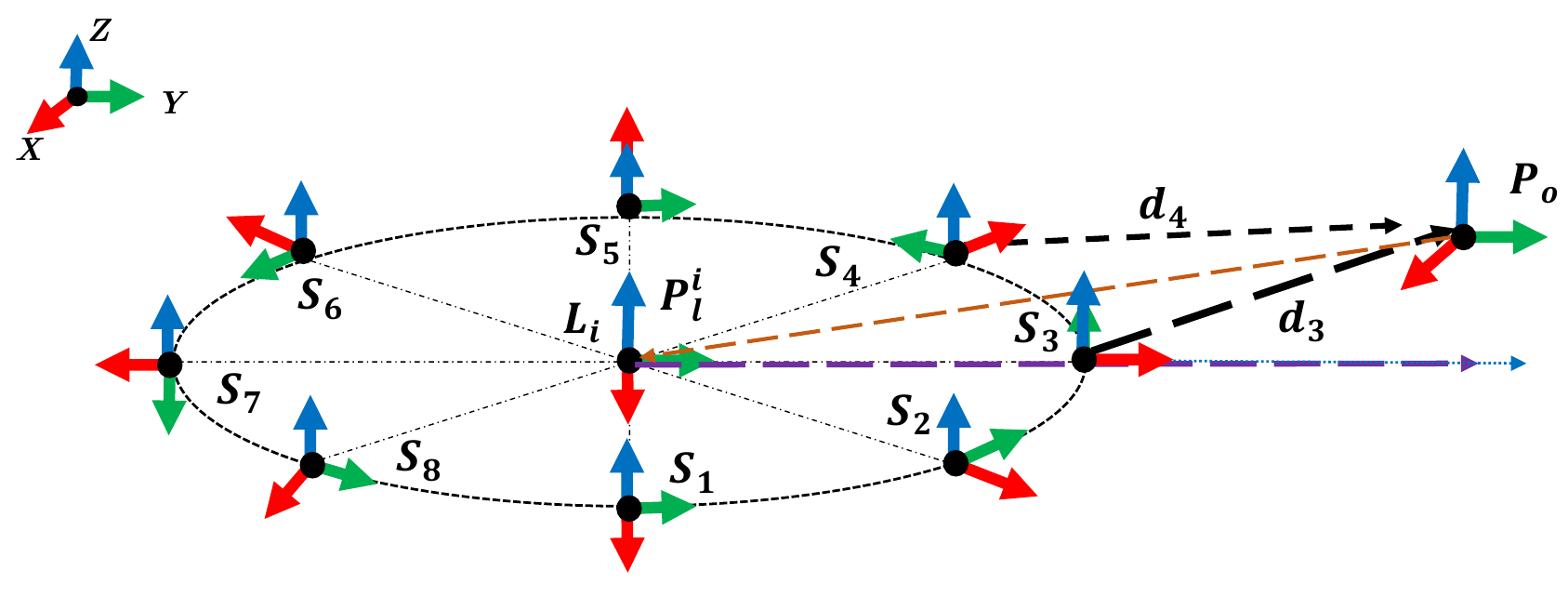}
	
\caption{The reference frames $(S_j : j=[1,8])$ placed at sensor locations in a ring fashion, to represent the transforms from the center of a link $l$.}
	\label{fig:sensor_tfs}
\end{figure}
 Then a minimum distance vector, from the center of the link $l_i$ to the obstacle is given as $\Vec{^{i}P_{lo}}$. Given a local transformation  with respect to link $l_i$, for the ring $i$, reported by sensor $j$,  $\Vec{D_{ij}}$ (see Figure \ref{fig:sensor_tfs}) can be defined as   
\begin{subequations}
\begin{equation}
 {d_{i_{min}}} = \min(d_1,d_2,d_3 \dots d_8) 
\end{equation}
\begin{equation}
    \Vec{D_{i}} = \Vec{^{i}P_{lo}} =  ^{l}T_{ij} \left[d_{i_{min}} \text{  } 0\text{  } 0\right]^T 
\end{equation}
 \begin{equation}
 {\lVert\Vec{D_i}}\rVert = ^{i}r_{l} + d_{i_{min}} 
\end{equation}
\begin{equation}
	{dsi_i} = \frac{1}{\lVert\Vec{D_i}\rVert ^2} = \frac{1}{(^{i}r_{l} + d_{i_{min}}) ^2}
\end{equation}
 \end{subequations}
where $d_{i_{min}}$ is the minimum distance measured by the TOF ring  $i = (base,elbow,tool)$, $\vec{P_{lo}}$ is the detected obstacle point w.r.t. the center of the link $(center,radius) \equiv (\Vec{^{i}P_{l}},^{i}r_{l}) $ of the TOF sensor ring, $^{l_i}T_{S_j}$ is the transformation matrix for the TOF sensor node $S_j, j=(1\dots8)$ to the center $^{i}P_{l}$ of the ToF sensor ring (see Figure \ref{fig:sensor_tfs}) and $dsi_i$ safety index for the TOF ring mounted on link $l_i$.

Here we ignore the uncertainty  in the position reading in the $yz$ axis, and compensate it in the $C_{dC}$ cushioning constant (Eq. 8b ). Similarly, the directed speed $^{i}k_o(t)$ can be equated as the minimum distance change with respect to the ring $i$ i.e.
\begin{equation}
^{i}k_o(t) = (\Vec{^{i}V_l}-\Vec{V_o}) \bullet \vec{^{i}P_{lo}} \Leftrightarrow \left(\frac{d_{i_{min}}(t) - d_{i_{min}}(t-\Delta t)}{\Delta t}\right)
\end{equation}

\subsubsection{\textbf{Self Occlusion Detection}}
\label{sec:SelfOcclusionDetection}
\begin{figure}[tb]

		\includegraphics[width =0.8\linewidth,keepaspectratio]{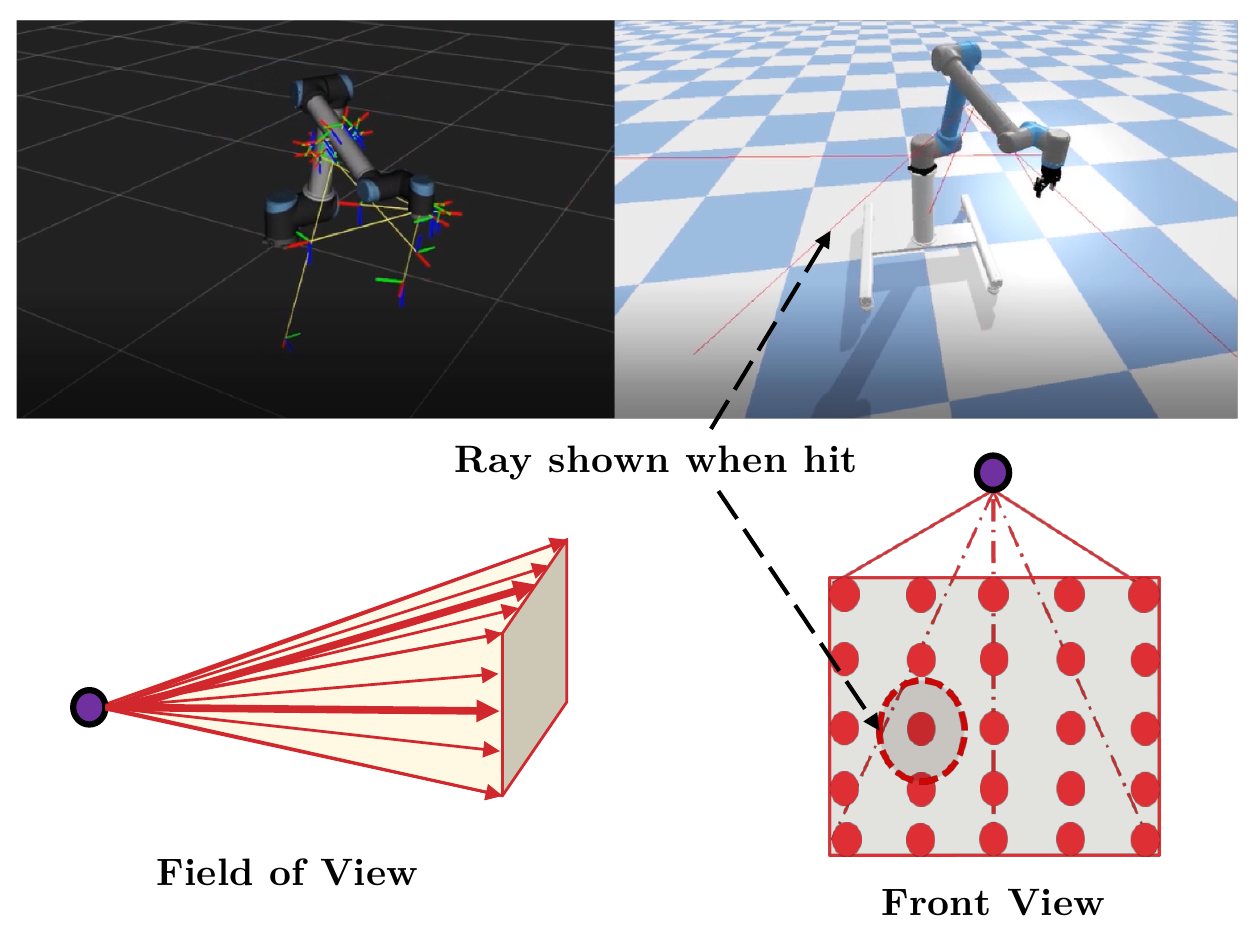}
			\centering

\caption{Self Occlusion Check using ray casting of all TOF sensors using PyBullet Physics Engine. The sensor is depicted as a point source, the lower half of the figure shows the perspective and front view of the FoV of the sensor.}
\label{fig:selfocclusioncheck}
\end{figure}

 The sensors mounted on each rings will also detect other robot links and objects in the work-space. In order to ignore the sensor readings, the sensor detection is modelled as shown in Fig. \ref{fig:selfocclusioncheck} using ray-casting in a physics engine \cite{pybullet2018}. Let the collection of objects that are stationary and belong to the restricted workspace of the robot be $W_{restricted}$ and that of the robot links or attached to the links be $W_{robot}$. For a sensor $j$ on TOF ring $i$, given a sensor measurement $d_{ij}$ and sensor accuracy $\sigma$, the ray-cast results in a distance $r_j$ of intersection of object $O_k$ in the workspace. The self-occlusion binary mask $m_{ij}$ can be written as follows:  
 \begin{subequations}
 \begin{equation}
 m_{ij}= 
\begin{cases}
    0,& \text{if}  \lbrace O_k \in (W_{restricted} \cup W_{robot} \rbrace \And r_j \in [d_{ij} \pm 3\sigma]  \\
    1,& \text{otherwise }  
\end{cases}
 \end{equation}
 \begin{equation}
 \hat{d_{ij}} = d_{ij} * m_{ij} 
 \end{equation}
 \end{subequations}
This is very similar to collision masking in a physics engine. The operation of ray-casting can be expensive if the rays of intersection are displayed, but if the process is running as an independent process, the average execution time taken is $\approx 2ms$ or $\ 500Hz$  for $600$ rays. This is enough to successfully mask the TOF rings sensors reading data at $30-50Hz$.   
%
\subsection{Robot Interface}
\label{sec:Robot Interface}
It is important to monitor and command a robot in real time to successfully implement SSM or any safety setup. Moreover, the robot desired speeds cannot be immediately achieved as it could generate large jerks and potentially cause damage to robot actuators.  The following sections provide information about the robot communication and transition of robot speeds to reduce jerk.


\subsubsection{\textbf{Robot Communication and Control}}
\label{sec:Robot_Comm_Control}
The robot used to test the SSM using the TOF rings is a Universal Robot UR10. The robot provides its internal state at 125Hz, through a robust communication protocol called RTDE. As the intended purpose of this application is not to control the robot motion but to change the speed, a `\textit{speed fraction}' value that behaves as explained in Eq. \ref{eq:jacobian_speed_fraction} is provided for control by UR10 interface. This is done by setting this value in real time using a TCP based real time interface. The robot provides the robot-state information at 125Hz and is assumed to be accurate. For this implementation the following information is used : joint variables $(q(t),\Dot{q(t)},\Ddot{q(t)})$, the robot TCP velocity $V_r(t)$ and  digital IOs pin states. 


\subsubsection{\textbf{Transitioning between robot collision states}}
\label{sec:transion_robot_coll_states}
In order to reduce jerks and smoothly transition between changing states, Reflexx Motion Library Type II \cite{krogerOpeningDoorNew2011} is used to generate the joint speed fraction $\rho$. This ensures that sudden robot state transitions does not abruptly change robot speeds. 

    




\section{Experiments and Validation}
\label{sec:experiment_validation}


\begin{figure*}
    \centering
    \includegraphics[width=1\textwidth]{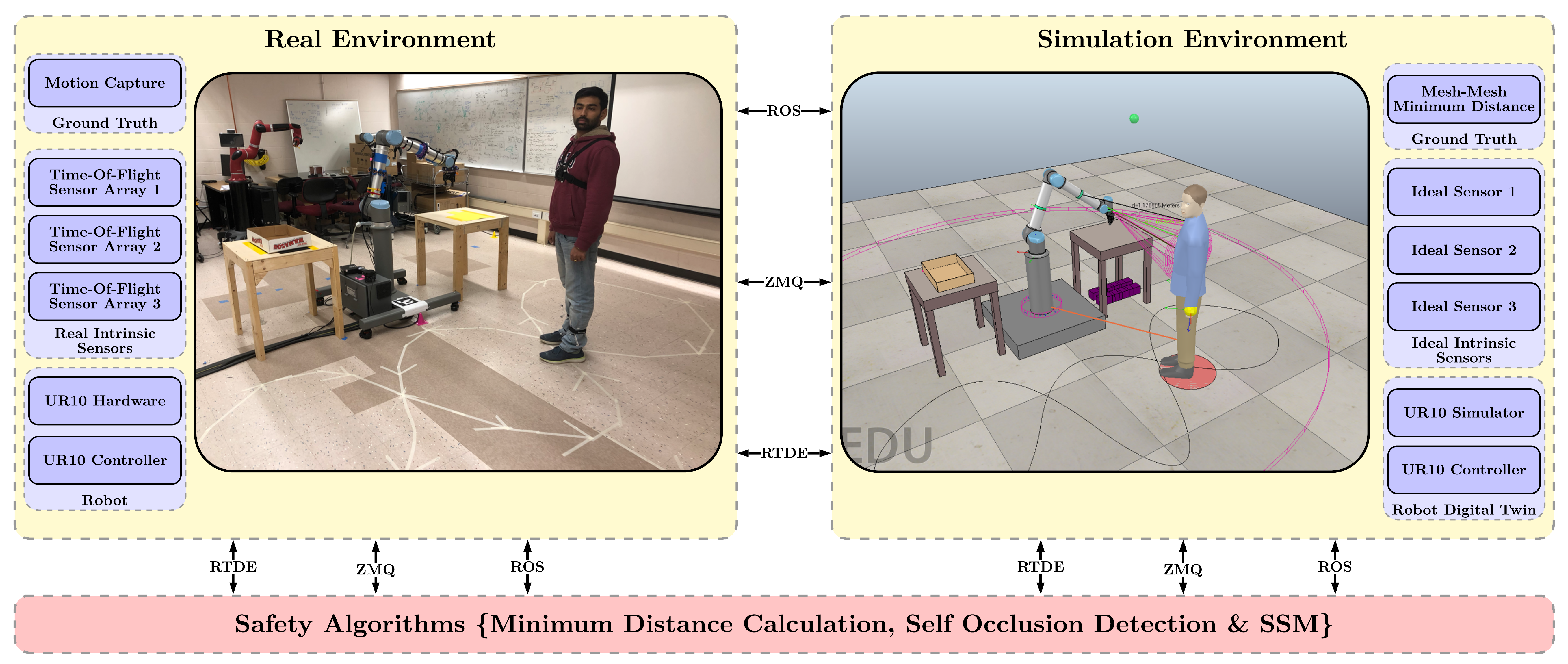}
    \caption{A schematic of the system used to implement, validate and test the proposed SSM safety configuration(s). The transport layer for communication between different subsystems such as the real and virtual environments is built using ZeroMQ, RTDE and ROS.}
    \label{fig:new-setup-diag}
\end{figure*}
The experiments performed the Safety, Performance and Productivity of  the Tri Modal SSM safety configurations (TriSSM-Vo,TriSSM-Vr \& TriSM) for minimum distance(s) calculated using the \textit{Real} ToF rings mounted on the robot, the \textit{Ideal} minimum distance from center of the links to the human/obstacle measured using motion capture and V-REP environment,  and 2D scanning \textit{Lidar}. A schematic of the experiment and validation setup is shown in Figure \ref{fig:new-setup-diag}. 

\subsection{Experiment Setup}
\label{sec:experiment setup}
The experiment setup is a generic robot pick and place task of placing 10 products in a box. The robot movement involves moving the base joint $180\deg$ between the pick and place positions on the tables (Figure \ref{fig:new-setup-diag}). This task was chosen as the base joint of a robot has the largest braking distance when moving at high joint speeds \cite{URSafetyFunc}. This results in a radial motion of the Tool-Control-Point(TCP) i.e. the end-effector.    

\subsection{Validation}
\label{sec:validation}
In order to validate and compare the SSM safety configurations for proximity based sensors, it requires the use of controlled physical avatars that can move in repeatable trajectories to maintain the same human interaction.
In this setup the human path is fixed as `{\rlap{$\infty \text{ }$}\llap{} $\infty$}' shaped path overlapping the robot's operating workspace as shown in  Figure \ref{fig:new-setup-diag}. The previous work \cite{kumarDynamicAwarenessIndustrial2018} can be referred for more details of the task and setup. In order to validate the system, the V-REP simulation environment  \cite{rohmerVREPVersatileScalable2013} is used to generate a \textit{digital twin} \cite{negri2017review} of the experiment where the a human-avatar mimics the motion of the human tracked using a motion-capture system. The robot pose and movements in V-REP are updated based on the reported states and movement of the robot in the real world. A similar approach has been used in \cite{safeeaMinimumDistanceCalculation2019}. In the experiment for a given SSM safety configuration the human motion, the states reported by the robot and performance of the SSM Safety Algorithms are recorded for evaluation. The human skeleton movements as reported by the motion capture are recorded. For the ideal and 2D liars, the robot is made to \textit{hallucinate} the presence of the human in the real world by moving the human avatar in V-REP according to the human recorded movements. This ensures the human interaction across all minimum distance calculation approaches the same. This is done at $\approx 120Hz-125Hz $ based on the information provided by the robot and the motion capture system.

It is to be noted that V-REP is used strictly for validation of the \textit{Real} ToF ring sensor hardware to compare to the ground truth and \textit{Ideal} minimum distance(s).


\subsection{Evaluation Metric}
 It is important to evaluate the accuracy of the minimum distance calculation and compare it with the ground truth $\vec{D_{gt}}$ as this provides the error in the minimum distance calculation. The metric of Safety, Performance and Productivity as shown in Table \ref{table:evaluation metric} is calculated for all SSM configurations (\textit{Vo,Vr,SM}) and minimum distance calculation approaches (\textit{Real,Ideal,Lidar}).  

The productivity and safety metrics for SSM were defined in \cite{marvelPerformanceMetricsSpeed2013} as :
\begin{subequations}
\begin{equation}
\label{eq:productivity}
    Productivity = \frac{t_{NoHRI}}{t_{HRI}}
\end{equation}
\begin{equation}
\label{eq:Safety Metric}
    Safety\hspace{2pt} Metric = \frac{\lVert \Vec{D_{gt}}\rVert^2}{\lVert \Vec{V_{tool}}\rVert}
\end{equation}
\end{subequations}
where $t_{NoHRI}$ and $t_{HRI}$ are the time taken to complete the given task with no human interactions and with human interaction respectively; $\Vec{D_{gt}}$ is the ground truth and $\Vec{V_{tool}}$ is the robot TCP velocity.

The minimum distance for 2D lidar is calculated from the base of the robot. As described in \cite{marvelImplementingSpeedSeparation2017}, the robots operating work-space offset needs to be subtracted to determine the minimum distance vector. This is a conservative assumption as it will be seen in the results below. In this experiment setup the robot operating workspace, i.e. the distance of the the TCP from the base of the robot when moving has a radius of $ r_{o} = 0.82m$. Hence the minimum distance for the 2D lidar, given a sensor measurement $d_{lidar}$ can be as approximated as :
\begin{equation}
\label{eq:lidar_Cdc}
    \lVert D_{lidar} \rVert =
    \begin{cases}
    d_{lidar} - r_o ,& \text{if }  d_{lidar} > r_o\\
     d_{lidar}, & \text{otherwise} \\
\end{cases}
\end{equation}


\begin{table}[h]
	\centering
	\caption{An evaluation metric for the proposed system.}
	\resizebox{0.475\textwidth}{!}{%
		\begin{tabular}{|l||l|l|l|}
			\hline 
			\textbf{\textsc{Criteria}}& \multicolumn{3}{c|}{\textbf{\textsc{ Min. Distance Calc. Approach \& SSM Config.}}} \\ 
			\hline \hline\hline
			& \textbf{Real*} & \textbf{Ideal**} & \textbf{Lidar 2D} \\ 
			\hline
			& \textbf{Vo,Vr,SM} & \textbf{Vo,Vr,SM} & \textbf{Vo,Vr,SM} \\ 
			\hline\hline 
			\textbf{Safety} 
			& \multicolumn{3}{l|}{ \textbullet Velocity Change at Stop Event} \\  
			& \multicolumn{3}{l|}{ \textbullet Average human-robot separation distance} \\ 
			& \multicolumn{3}{l|}{ \textbullet Safety Metric as per Eq. \ref{eq:Safety Metric}} \\ 
			\hline 
			\textbf{Performance}
			& \multicolumn{3}{l|}{\textbullet Average Stopping \& Reduced Time} \\ 
			& \multicolumn{3}{l|}{\textbullet Average Reaction Time} \\   
			\hline 
			\textbf{Productivity} & \multicolumn{3}{l|}{\textbullet Time taken to complete the task}  \\
			& \multicolumn{3}{l|}{with HRI as per Eq. \ref{eq:productivity}} \\ 
			\hline 
			\multicolumn{3}{l}{\textit{*   Based on distances reported by ToF Rings sensor-hardware}}\\
			\multicolumn{3}{l}{\textit{**  Based on Mocap \& V-REP based Mesh-Mesh distance}}\\
		\end{tabular} 
	}
	\label{table:evaluation metric}
\end{table}
\section{Results}
\label{sec:Results}
The experiments were performed with the parameters given in Table \ref{table:Parameters Used}. There were five trials for each experiment i.e. in total results from 45 experiments tabulated.
\begin{table}[h]
\caption{Parameters used for the SSM config. for different Min.Distance Calculation Approach}
\label{table:Parameters Used}
\resizebox{0.475\textwidth}{!}{%
\begin{tabular}{|l|l|l|l|}
\hline
 & Real & Ideal & Lidar \\ \hline
\multirow{3}{*}{\begin{tabular}[c]{@{}l@{}}TriSSM-Vo\\ TriSSM-Vr\end{tabular}} & \multicolumn{3}{l|}{$T_{red}=0.4s$ ,$T_{stop}=0.4s$ , $T_{r}=0.1s$} \\ \cline{2-4} 
 & \multicolumn{3}{l|}{*$V_{lmax}=1.7m/s $,$**V_{hmax}=1.6m/s$, $k_{max}=V_{lmax}+V_{hmax}=3.3m/s$} \\ \cline{2-4} 
 & \multicolumn{2}{l|}{$C_{dC}=0.3m$, $B_{min} = 0.2m$} & $r_o=0.82m$, $C_{dC} = r_o + 0.3m = 1.12m$ \\ \hline
Tri-SM & \multicolumn{2}{l|}{$\lbrace d_C,d_R \rbrace = \lbrace 0.5m,1.1m \rbrace $} & $\lbrace d_C + r_o ,d_R + r_o \rbrace = \lbrace 1.32m,1.92m \rbrace$ \\ \hline
			\multicolumn{2}{l}{\textit{* max. TCP velocity for the given task.}}\\
			\multicolumn{2}{l}{\textit{** max. Human walking speed \cite{marvelImplementingSpeedSeparation2017}}}
\end{tabular}%
}
\vfill
\end{table}

\begin{figure}
    \centering
    \begin{flushright}
        \includegraphics[clip, trim=6.1cm 22.665cm 1.3cm 4.2cm, width=0.475\textwidth]{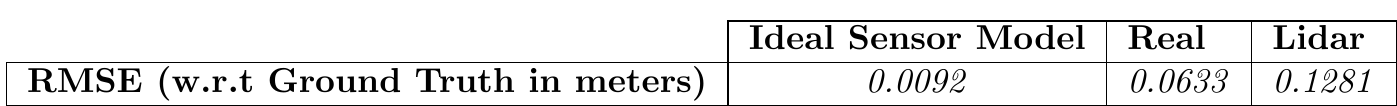}
        \end{flushright}
        \vspace{-5pt}
        \includegraphics[clip, trim=5.00cm 19cm 2.75cm 4.5cm, width=0.5\textwidth]{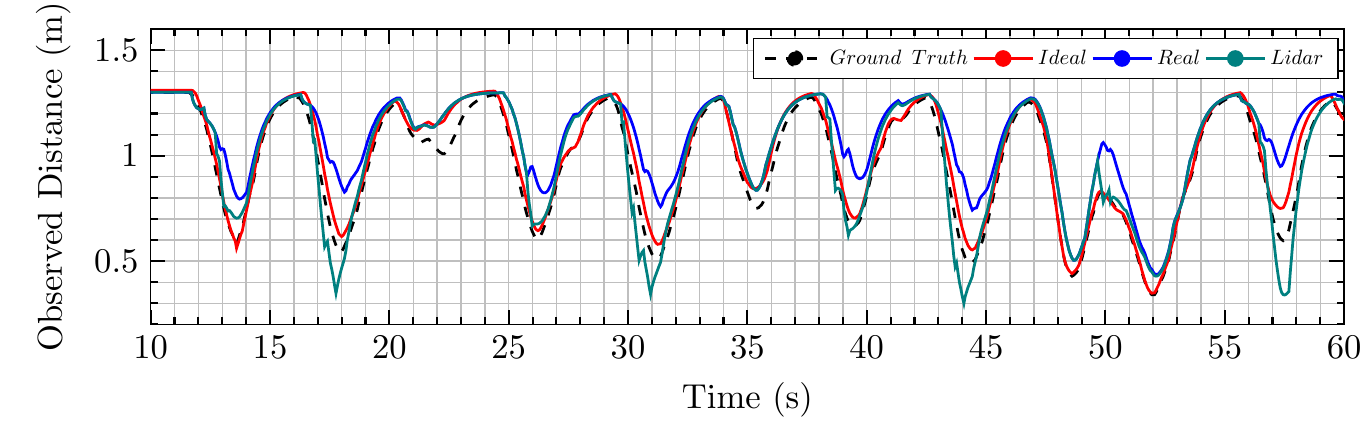}
        \caption{Minimum Distance Calculation Comparison w.r.t. the ground truth in Terms of RMSE (meters)}
    \label{fig:minimum_distance_graph}
\end{figure}

Figure \ref{fig:minimum_distance_graph} shows the comparison of the minimum distance calculated w.r.t. the ground truth, $\lVert\Vec{D_{gt}}\rVert$. The RMSE of \textit{Ideal} minimum distance $\lVert\Vec{D_{ideal}}\rVert$ is approximately $9mm$, which validates our assumptions of $\lVert\Vec{D_{ideal}}\rVert \approx \lVert\Vec{D_{gt}}\rVert$ for this experiment setup.  The RMSE of the  \textit{Ideal} minimum distance from the ToF sensor rings is $63mm$ and \textit{Lidar} is $128mm$. The RMSE of \textit{Lidar} is larger because it reports the distance from the fixed base of the robot and its visibility is limited to measuring the position of the human legs or the lower torso.

The RMSE of the ToF sensor is larger than the ideal sensor because of the sensor reading accuracy and the sparsity of the sensor nodes (8), that can create blind-spots and abrupt jumps in the range reading(s). This sensor detection uncertainty is compensated in the $C_{dc}$ cushioning constant as $Z_s$ as described in Eq. \ref{eq:ssm_linear}. Therefore, it can be concluded that ToF sensor rings can be used for implementing SSM. 

\begin{figure}
\centering
\fbox{\includegraphics[clip, trim=3cm 7.4cm 3.5cm 3cm, width=0.45\textwidth]{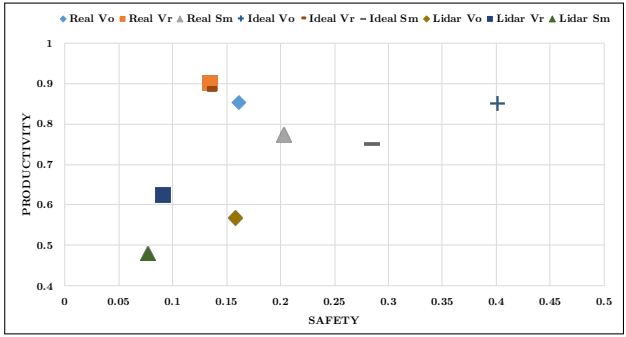}}
\caption{Productivity vs Safety Metric graph; for all SSM safety configurations, for all minimum distance calculation approaches.}
\label{fig:Productivity_Safety}
\end{figure}

According to the Safety formulation in Eq. \ref{eq:Safety Metric}, the safest SSM configuration is the one where the robot doesn't move. But that also means there is no productivity. Hence, the `\textit{usably safe}' SSM configuration that optimizes the safety and productivity should be preferred \cite{marvelPerformanceMetricsSpeed2013}. The graph shown in Figure \ref{fig:Productivity_Safety} plots productivity against safety during a task performed with a given SSM safety configuration for all sensors. It is observed that the \textit{Real} and \textit{Ideal} based minimum distance have higher productivity than \textit{Lidar}. The highest productivity is given by TriSSM-Vr for \textit{Real} and \textit{Ideal}, however this configuration is less safe. The highest safety is reported for TriSSM-Vo for \textit{Ideal} minimum distance. It can be further seen, that for a given minimum distance calculation TriSSM-Vo is safer than TriSM-Vr.

The \textit{Lidar} based SSM reports lower safety levels. This is because the Lidar need to consider a larger intrusion distance human/obstacle $C_{lidar} = C + r_o + r_h $, compared to the smaller value considered for the Real and Ideal distances, refer Eq. \ref{eq:ssm_linear}and Eq. \ref{eq:lidar_Cdc} \cite{marvelImplementingSpeedSeparation2017}. In order to keep the comparison same, for a given $C$ , the \textit{Lidar} based TriSSM safety configuration behaves relatively less safe in comparison to \textit{Real} and\textit{ Ideal}. Increasing the intrusion distance makes the SSM configuration safer at the cost of decreased productivity.
As Safety metric is obtained from the ground truth and the robot TCP velocity $V_{tool}$, the \textit{Real} sensors are less safe due to the error in minimum distance calculation.

Thus it can be observed that given the minimum distance calculation is accurate, a safer and more productive SSM configuration can be implemented by using on-robot sensors. Also the consideration of robot and human/obstacle directed speeds in SSM i.e. TriSSM-Vo adds to the safety and productivity. Using robot directed speed alone result in less safe SSM configuration.

 For ease of visualization and comparison, the separation distances, velocity changes, reaction times and robot times to stop and reduce, are plotted as radar graphs as shown in Figure \ref{fig:radar_graphs}. It can be observed in Figure \ref{fig:radar_graphs}(a) that the fastest reaction is of the TriSSM-Vo for the \textit{Real} sensor setup. The reaction times are negative to denote that the robot anticipated a stop event before the minimum distance reached the critical distance threshold. The average reaction time of the system represents the sensitivity and responsiveness to distance and directed speed changes. The \textit{Real} based SSM are more responsive as the time taken to determine the minimum distance is faster than the calculation of the \textit{Ideal} distance.
 
 The average stopping and reduce times indicate the anticipatory nature to human/obstacle motion in the shared workspace. As seen in Figure \ref{fig:radar_graphs}(b), the time to stop is higher for all Tri-SM and the time to reduce speed is higher for TriSSM-Vr and Tri-SM. It can be observed that TriSSM-Vo gives the best results. 
 
 The `velocity before stop' graph , Figure \ref{fig:radar_graphs}(c) represents how often there are sharp deceleration(s) at the stopping event. It can be seen that TriSSM-Vr has the highest velocity change. This can cause more wear on the actuators and sudden speed changes can be uncomfortable for the human sharing the workspace.
 
 The separation distance gives an idea of when reduced or stop events were triggered and what was the average separation distance. SSM based on \textit{Real} and \textit{Ideal} minimum distances  have nominal stop and reduce distances averaged around $0.5m$ and $0.75m$. However, the \textit{Lidar} is more conservative. 
\begin{figure*}
	\centering
	\includegraphics[clip, trim=5cm 17.5cm 0.5cm 4.5cm, width=0.9\textwidth]{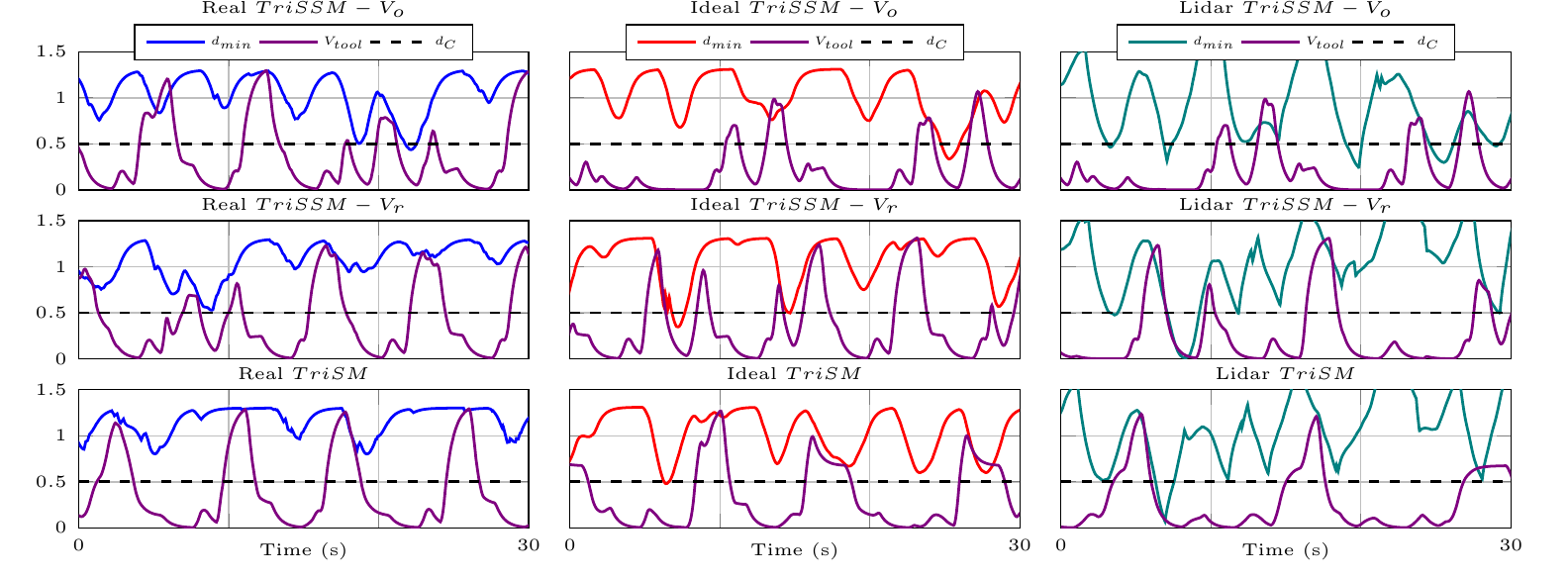}
	\caption{ The response of the robot TCP velocity to the change in the measured minimum distance. The columns represent the minimum distance measured using - ToF rings, Ideal distance and Lidar, and the rows are the SSM safety configurations - TriSSM-Vo, TriSSM-Vr and TriSM.}
	\label{fig:distvelgraph_big}
\end{figure*}
Lastly, Figure \ref{fig:distvelgraph_big} shows the tool velocity to the distance measured for all SSM and minimum distance configuration(s). The graph depicts the response of the robot TCP velocity to the change in the measured minimum distance.




 

\begin{figure}
\centering
\includegraphics[width=0.22\textwidth]{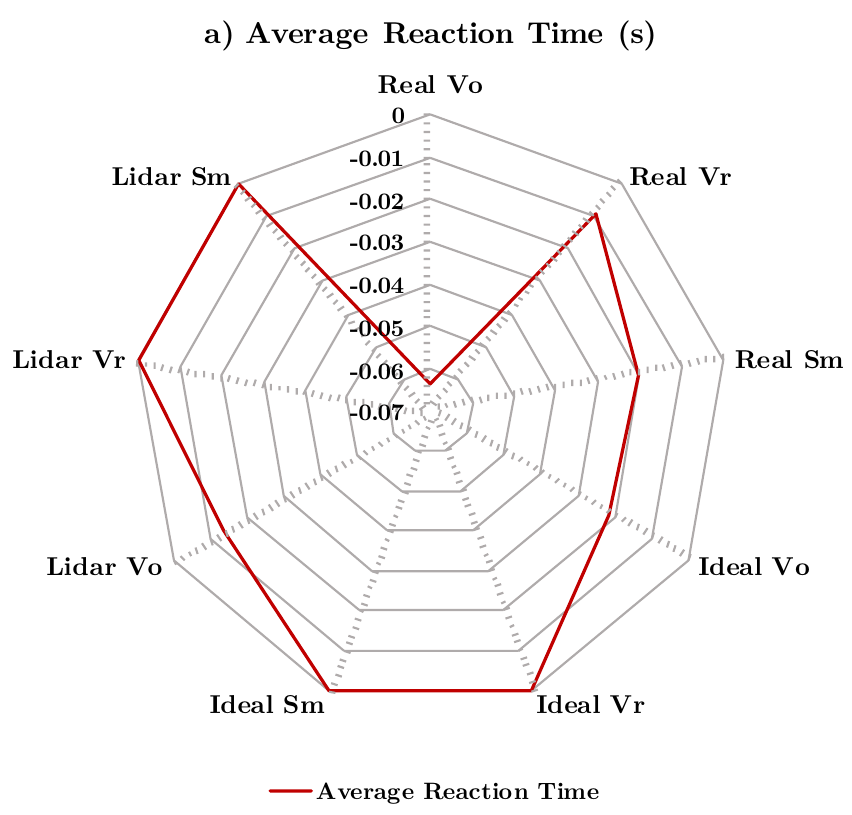}
\includegraphics[width=0.22\textwidth]{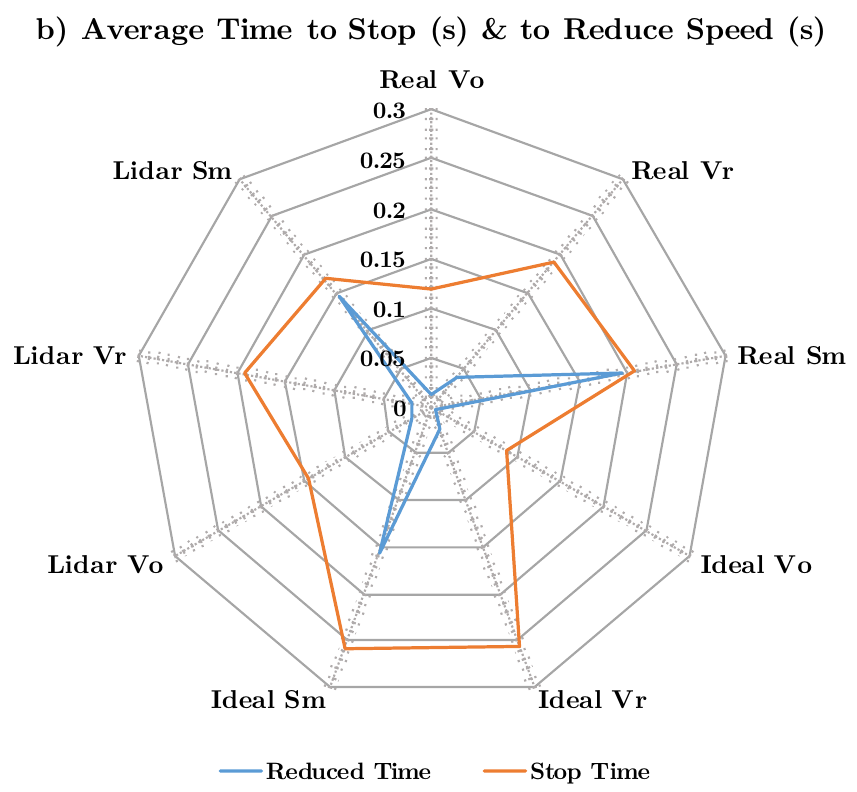}
\includegraphics[width=0.22\textwidth]{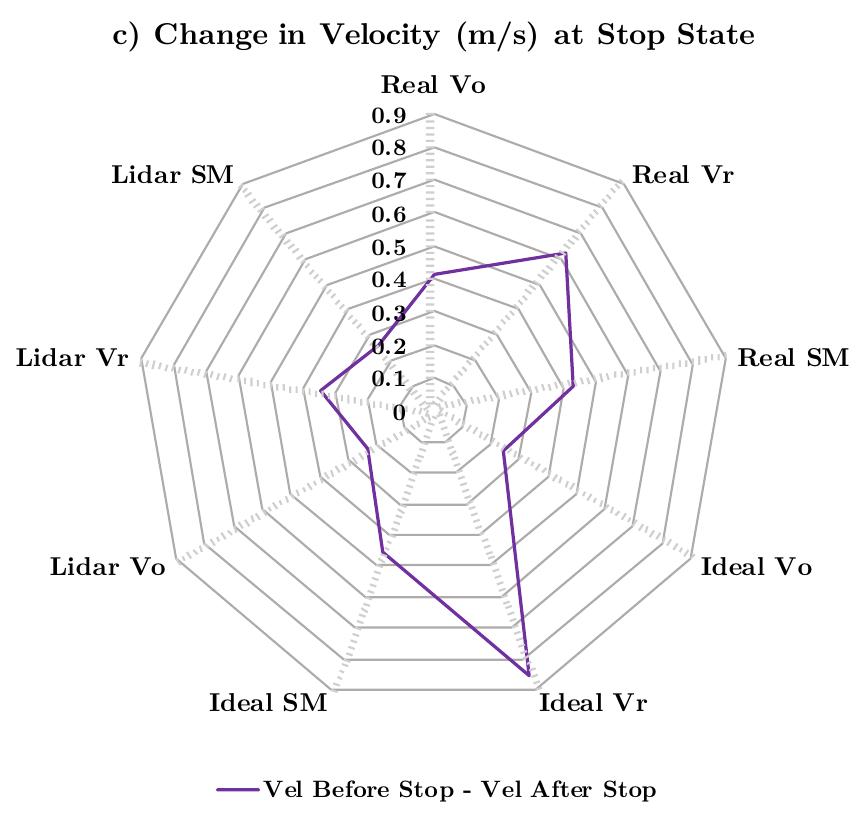}
\includegraphics[width=0.22\textwidth]{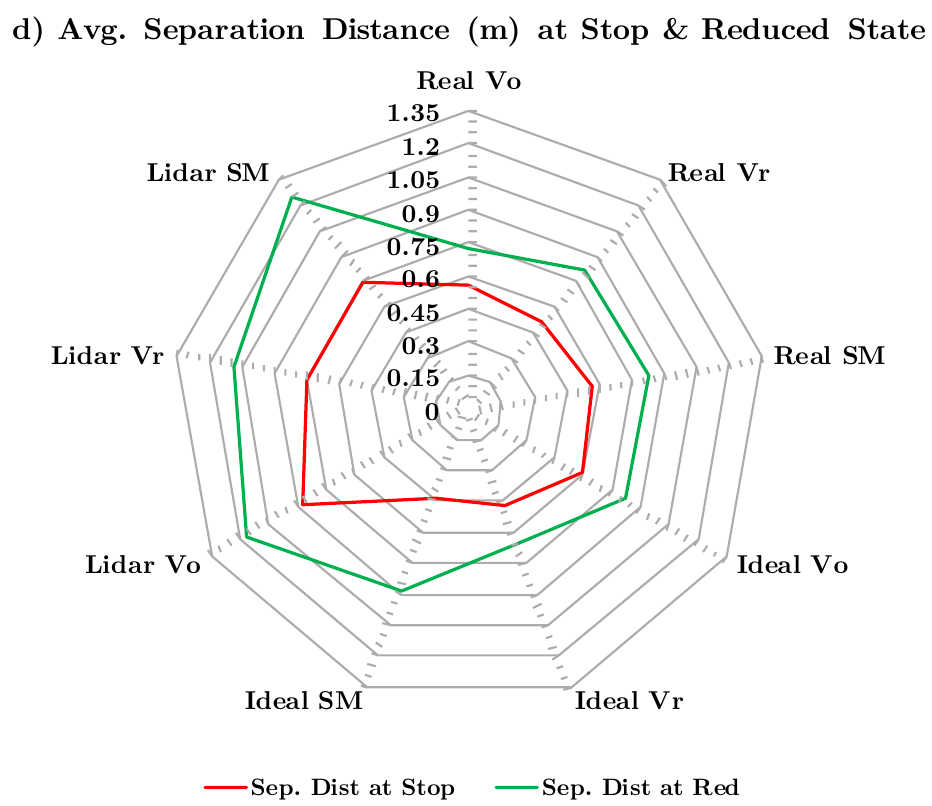}
\caption{Radar graphs comparing a) Average Reaction Time of the system b) Average Time to Stop and Reduce c) The change in velocity at a Stop event d) Average human/obstacle robot separation distance at reduce and stop events; for all SSM safety configurations, for all minimum distance calculation approaches.}
\label{fig:radar_graphs}
\end{figure}






The video demonstrating the experiment setup and the performance of the ToF rings can be viewed at \href{https://drive.google.com/open?id=1m2AmNXJJX42zNEc8Bn6rjPUtgPPUTZP7}{link}. \footnote{http://tinyurl.com/case2019-tofssm}



\section{Conclusion}
\label{sec:Conclusion}
In this paper, a Time-of-Flight sensor ring as a form of on-robot sensing for speed and separation monitoring is proposed . From the simulation and real-world implementation results so far there is a significant benefit in terms of safety, performance  and productivity  during an HRC task in comparison to conventionally used 2D scanning lidars. The true minimum distance between human/obstacle and a robot can be approximated with measuring minimum distance from centers of the robot links. The Tri-Modal SSM that leverages both relative human-robot speeds and separation distance result in more consistent and smoother robot movement. The results of this work are intended to provide the design for a simple plug and play device as an alternative or addition to current 2D Lidar scanners for optimizing productivity while ensuring human safety.

It was observed that if a minimum distance calculation is accurate, a safer and more productive SSM configuration can be implemented by using on-robot ToF sensors. A prototype addressing the issues of sensor uncertainty due to blind spot has been made and tested. It has 16 TOF sensor nodes compared to the 8 used here, and has smaller footprint.







\section*{Acknowledgment}
The authors are grateful to the staff of Multi Agent Bio-Robotics Laboratory (MABL) and the CM Collaborative Robotics Research (CMCR) Lab for their valuable inputs.


\bibliographystyle{IEEEtran}
\bibliography{CASE2019_TOF}{}

\end{document}